\documentclass[conference]{IEEEtran}
\IEEEoverridecommandlockouts
\usepackage{cite}
\usepackage{amsmath,amssymb,amsfonts}
\usepackage{algorithmic}
\usepackage{graphicx}
\usepackage{textcomp}
\usepackage{xcolor}
\usepackage[ruled,vlined,lined,linesnumbered]{algorithm2e}
\usepackage[caption=false,font=normalsize,labelfont=sf,textfont=sf]{subfig}
\usepackage{amsmath}
\usepackage{amssymb}
\newcommand{\methodname}{{\tt{FedDua}}}
\def\BibTeX{{\rm B\kern-.05em{\sc i\kern-.025em b}\kern-.08em
    T\kern-.1667em\lower.7ex\hbox{E}\kern-.125emX}}
\begin{document}

\title{Local Data Quantity-Aware Weighted Averaging for Federated Learning with Dishonest Clients}

\author{
    \textit{Leming Wu}$^{1}$, \textit{Yaochu Jin}$^{1,2,*}$\thanks{*Corresponding author.}, \textit{Kuangrong Hao}$^{1}$, \textit{Han Yu}$^{3}$ \\ \\
    {$^1$College of Information Science and Technology, Donghua University, Shanghai, China} \\
    {$^2$School of Engineering, Westlake University, Hangzhou, China} \\
    {$^3$College of Computing and Data Science, Nanyang Technological University (NTU), Singapore} \\
    {\small \texttt{lemingwu@mail.dhu.edu.cn, jinyaochu@westlake.edu.cn, krhao@dhu.edu.cn, han.yu@ntu.edu.sg}}
}

\date{}

\maketitle

\begin{abstract}
Federated learning (FL) enables collaborative training of deep learning models without requiring data to leave local clients, thereby preserving client privacy. The aggregation process on the server plays a critical role in the performance of the resulting FL model. The most commonly used aggregation method is weighted averaging based on the amount of data from each client, which is thought to reflect each client's contribution. However, this method is prone to model bias, as dishonest clients might report inaccurate training data volumes to the server, which is hard to verify. To address this issue, we propose a novel secure \underline{Fed}erated \underline{D}ata q\underline{u}antity-\underline{a}ware weighted averaging method (\methodname{}). It enables FL servers to accurately predict the amount of training data from each client based on their local model gradients uploaded. Furthermore, it can be seamlessly integrated into any FL algorithms that involve server-side model aggregation. Extensive experiments on three benchmarking datasets demonstrate that \methodname{} improves the global model performance by an average of 3.17\% compared to four popular FL aggregation methods in the presence of inaccurate client data volume declarations.
\end{abstract}
\section{Introduction}
\label{sec:intro}

In recent years, artificial intelligence (AI) technology has made significant progress and is now integrated into many aspects of daily life, including intelligent question-answering systems \cite{shanahan2023role}, smart finance \cite{singh2023privacy}, and autonomous driving \cite{chen2024end}, etc. While the development of AI has greatly benefited society by providing convenience, it has also raised concerns about the leakage of user privacy data, as technology can act as a double-edged sword. Federated learning (FL) \cite{goebel2023trustworthy,jin2023federated,fan2025ten}, which allows deep learning models to be trained without transferring local data, can effectively mitigate the risk of privacy breaches.

FL can be classified into centralized FL \cite{mcmahan2017communication, wu2024fl}, which involves a central node server, and decentralized FL \cite{beltran2023decentralized}, which does not. Additionally, based on the relationship between data labels and features, FL can be categorized into horizontal \cite{yang2020federated, lyu2024privacy} and vertical types \cite{ren2025advances}. This paper primarily focuses on centralized FL with a central node server. In this setup, the server typically performs weighted aggregation of model parameters. As shown in Eq. \eqref{fl_aggregation}, the most common method of weighted aggregation is based on the amount of client data. This approach has several advantages in FL. It better reflects the contribution of data to the global model, improves the model's generalization ability, and reduces the impact of noise from clients with small sample sizes, while balancing fairness and efficiency. In addition, this method is simple, easy to implement, and easy to interpret.

\begin{figure}[htbp]
	\centering
    \subfloat{\includegraphics[width=.5\linewidth]{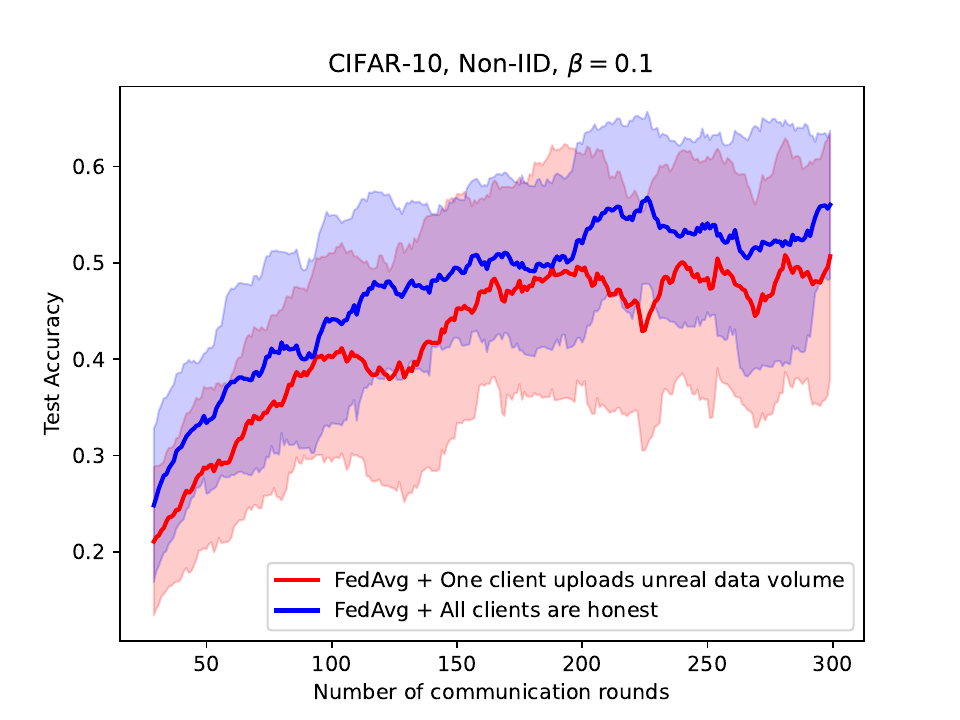}} 
    \subfloat{\includegraphics[width=.5\linewidth]{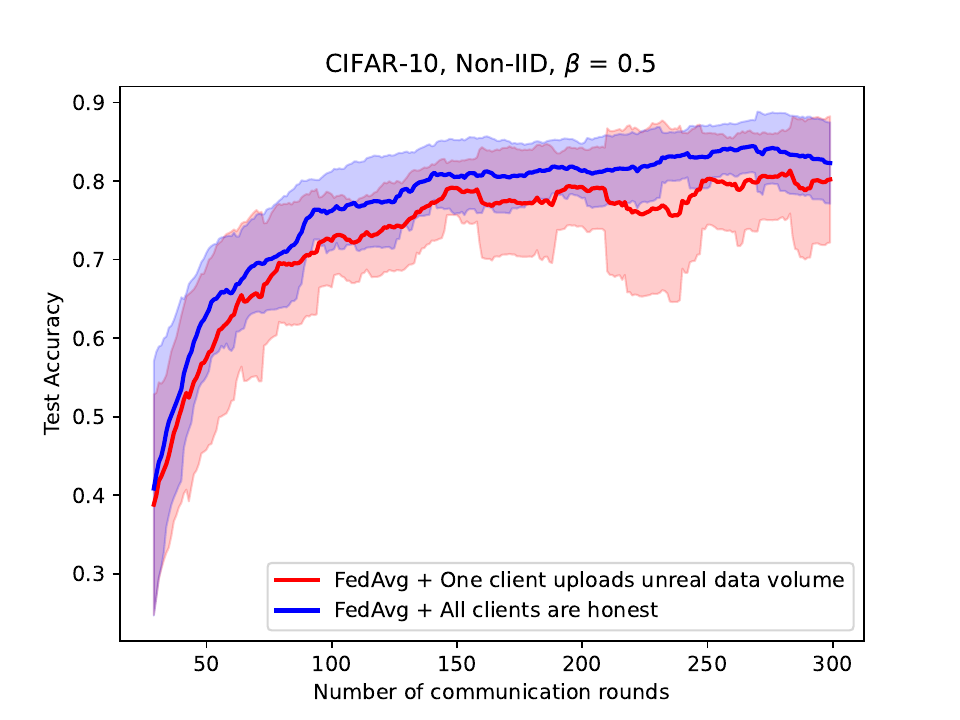}}
	\caption{This experiment evaluates the global model accuracy of the FedAvg algorithm on the CIFAR-10 dataset, comparing scenarios with and without dishonest clients manipulating data volume. Two levels of Non-IID data distribution, defined by Dirichlet parameters $\beta = 0.1$ and $\beta = 0.5$, are considered.}
    \label{fig1:motivation}
\end{figure}

When performing weighted aggregation based on the amount of client data, the server can only aggregate according to the data reported by the clients. Dishonest clients can manipulate their local model weights by providing false data amounts. We experimentally verified this behavior using the classic FL algorithm, FedAvg. In each round, 10 clients were randomly selected from a total of 100 to participate in federated training. If one client dishonestly reports its data amount, we observed a significant drop in the global model's accuracy, as shown in Fig. \ref{fig1:motivation}.

To address this issue, we propose a first-of-its-kind secure \underline{Fed}erated \underline{D}ata q\underline{u}antity-\underline{a}ware weighted averaging method (\methodname{}). The core idea is to design a data quantity-aware branch and integrated it into the client model. This branch accurately predicts the adjustment factor $\alpha$ related to the amount of data. By combining $\Delta\theta$ from the client model, the learning rate, and the average gradient from client training, we can estimate the client's data amount. Additionally, we found that the distribution of the adjustment factor $\alpha$ remains similar across different amounts of data. Therefore, once the server receives the adjustment factor $\alpha$ from the client, it can verify the reported data amount by comparing it with the pre-trained distribution of $\alpha$. If a dishonest client is detected, the server will issue a warning, encouraging the client to report the data amount honestly. If the client persists in dishonesty, the server can exclude it from federated training.

Our proposed \methodname{} method can be implemented as a module in other federated algorithms that require aggregation. Extensive experiments have demonstrated that the algorithm can successfully identify clients that dishonestly report data amounts and mitigate the reduction in model accuracy caused by such dishonest reporting. Extensive experiments on three benchmarking datasets demonstrate that \methodname{} improves the global model performance by an average of 3.17\% compared to four popular FL aggregation methods in the presence of inaccurate client data volume declarations.

\section{Related Work}
Existing methods \cite{mcmahan2017communication, li2020federated, li2021ditto, karimireddy2020scaffold} for weighted aggregation in FL typically perform aggregation based on the amount of client data. This approach enhances the stability and generalization performance of the global model, making it an important and effective aggregation strategy. However, if a dishonest client uploads an incorrect amount of training data, the global model's performance will degrade. To address this, \cite{lyu2020collaborative, chen2020dealing} proposed a weighted aggregation method based on client contributions. However, methods that more accurately measure client contributions, such as those in \cite{wang2019measure, shi2023fairness}, often involve high computational complexity. \cite{yi2022qsfl} proposed using the client's verification loss as a measure of contribution, allowing weighted aggregation based on verification loss. However, this method is sensitive to data volume differences and Non-IID data, and it introduces additional computational overhead and instability. \cite{pillutla2022robust, li2023revisiting} proposed robust aggregation strategies, but these still rely on client data volume for weighted aggregation. 

While these methods improve the robustness of FL to some extent, they significantly reduce training performance when dishonest clients are present. The proposed \methodname{} approach is designed to bridge this important gap in current FL literature.

\section{Preliminaries}
In an FL scenario with a central server, there are $N$ clients, each with data $D_i=(x_i, y_i), i=\{1,…, \left| D_i \right|\}$, where $\left| D_i \right|$ denotes the number of data points on client $i$, and $(x_i, y_i)$ represents the input sample $x_i$ and its corresponding label $y_i$. In each communication round, the server selects $K$ clients to participate in federated training. Initially, the server sends the global model $\theta_g^0$ to each client. Upon receiving the model, client $i$ trains its local model $\theta_i^0$ using its local data $D_i$ without transferring data, as shown in Eq. \eqref{fl_obj}. Here, $f(x_i;\theta)$ is the model's prediction, and $l$ is the loss function. The client then uploads its model parameters $\theta_i^1$ to the server, where the parameters are aggregated, as shown in Eq. \eqref{fl_aggregation}.
\begin{align}\label{fl_obj}
\mathcal L_i(\theta_i;D_i) = \mathbb{E}_{(x_i, y_i) \sim {D}_i} \left[ l \left( f(x_i; \mathbf{\theta}), y_i \right) \right],
\end{align}
\begin{align}\label{fl_aggregation}
\min_{\{\{\theta_i\}\}_{i=1}^K} \frac{\left| D_i \right|}{\sum_{i=1}^K \left| D_i \right|} \sum_{i=1}^K \mathcal L_i(\theta_i;D_i).
\end{align}

If there is an incorrect data volume in the weighted aggregation of Eq. \eqref{fl_aggregation}, the actual data volume is denoted as $\left| D_j \right|_{real}$, and the reported (false) data volume is $\left| D_j \right|_{unreal}$, where $\left| D_j \right|_{real} \neq \left| D_j \right|_{unreal}$. The weight in the global loss function is determined by the data volume. If client $j$ dishonestly reports its data volume, the optimization objective becomes:
\begin{align}\label{fl_aggregation_}
\mathcal{L}^{'} = \sum_{i \neq j} w_i \cdot \mathcal{L}_i + \left( w_j + \Delta w_j \right) \cdot \mathcal{L}_j.
\end{align}
Here, $\Delta w_j=w_j^{'}-w_j$, which shows that if $\Delta w_i > 0$, the increase in $w_j+\Delta w_j$ amplifies the impact of client $j$ on the global model. Additionally, if the loss distribution $\mathcal L_j$ differs significantly from that of other clients, the global optimization direction will be biased towards client $j$.


\begin{algorithm}
\normalsize
 \SetAlgoLined
 \SetKwData{Left}{left}\SetKwData{This}{this}\SetKwData{Up}{up}
 \SetKwRepeat{doWhile}{do}{while}
 \SetKwFunction{Union}{Union}\SetKwFunction{FindCompress}{FindCompress}
 \SetKwInOut{Input}{Input}\SetKwInOut{Output}{Output}
 \For{each round of local model update}{
 S $\leftarrow$  The set of clients;\\
\For{each client $i$ in the set of clients $S$}{
$\theta_i^t$ $\leftarrow$ Train the local model by the local data in $i$-th client;\\
$\nabla\theta_i^t$ $\leftarrow$  Calculate the model's gradient in the local training;\\
$\alpha_i^t$ $\leftarrow$  Calculate the $\alpha$ by training data quantity-aware branch;\\
Send $(\nabla\theta_i^t,\alpha_i^t)$ to the server;\\
}
\textbf{On the server:} \\
Verify $\alpha$ and the amount of data on the client through $\alpha$, $\nabla\theta_i^t$ and $\Delta\theta$;\\
The server evaluates the reliability of the data provided by each client by combining $\alpha_i^t$ with the pre-trained distribution of $\alpha$;\\
\eIf{$\alpha_i^t$ is not True}{
$|D_i|$ $\leftarrow$ For clients that are found to upload unrealistic amounts of data, aggregate based on the predicted amount of data: $|D_i| = \frac{\Delta \theta_i}{\eta \cdot \nabla \theta_i \cdot \alpha}$
;\\
}{
$\theta_g^t$ $\leftarrow$ Aggregate based on the amount of training data uploaded by the client: $\theta_g^t = \sum_{i=1}^K \frac{|D_i|}{\sum_{i=1}^K |D_i|} \theta_i^t$
;\\
}
Send $\theta_g^t$ to clients in set $S$;
}
\caption{The pseudo-code of \methodname{}}
\label{alg1}
\end{algorithm}

\begin{figure*}[htbp]
    \centering
    \includegraphics[width=7in]{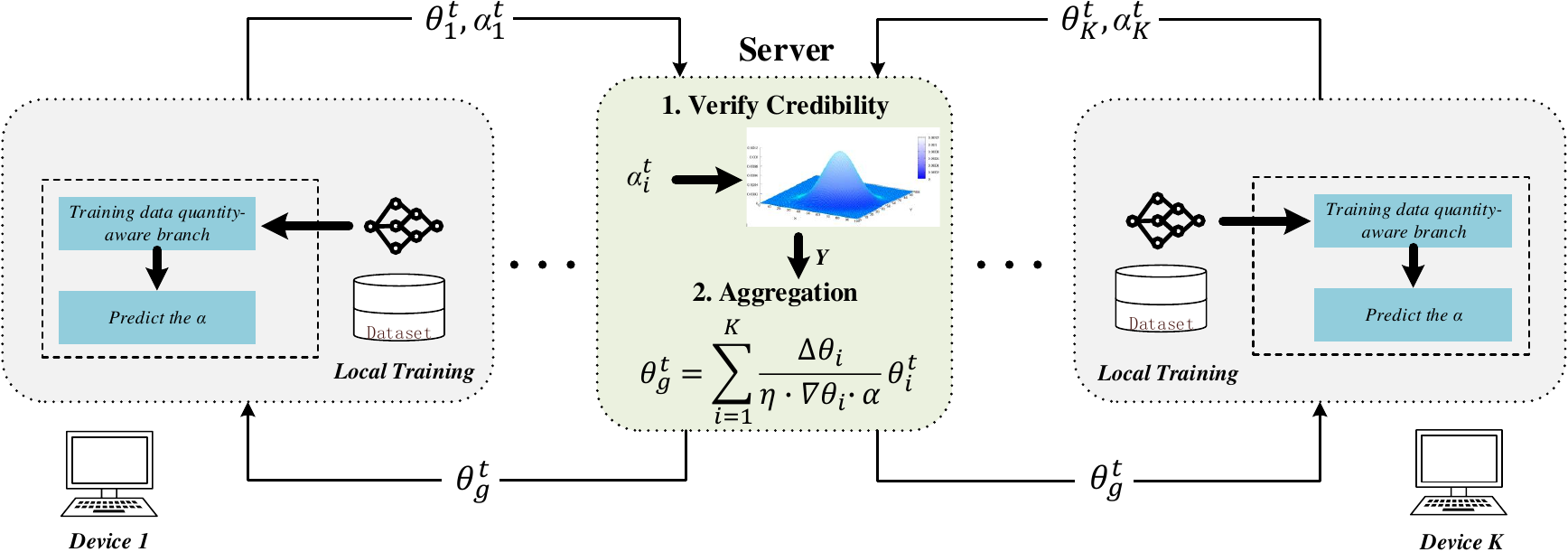}
    \caption{The proposed \methodname{} approach architecture.}
    \label{fig:Architecture}
\end{figure*}

\section{The proposed \methodname{} approach}
In this section, we describe the proposed algorithm \methodname{}. The algorithm consists of two main parts: the client obtains the adjustment factor $\alpha$ through the quantity-aware branch, and the server determines whether the data volume uploaded by the client is credible. The architecture of the proposed algorithm is shown in Fig. \ref{fig:Architecture}. The detailed process of the algorithm is provided in Algorithm \ref{alg1}.

\begin{figure*}[htbp]
	\centering
	\subfloat{\includegraphics[width=.25\linewidth]{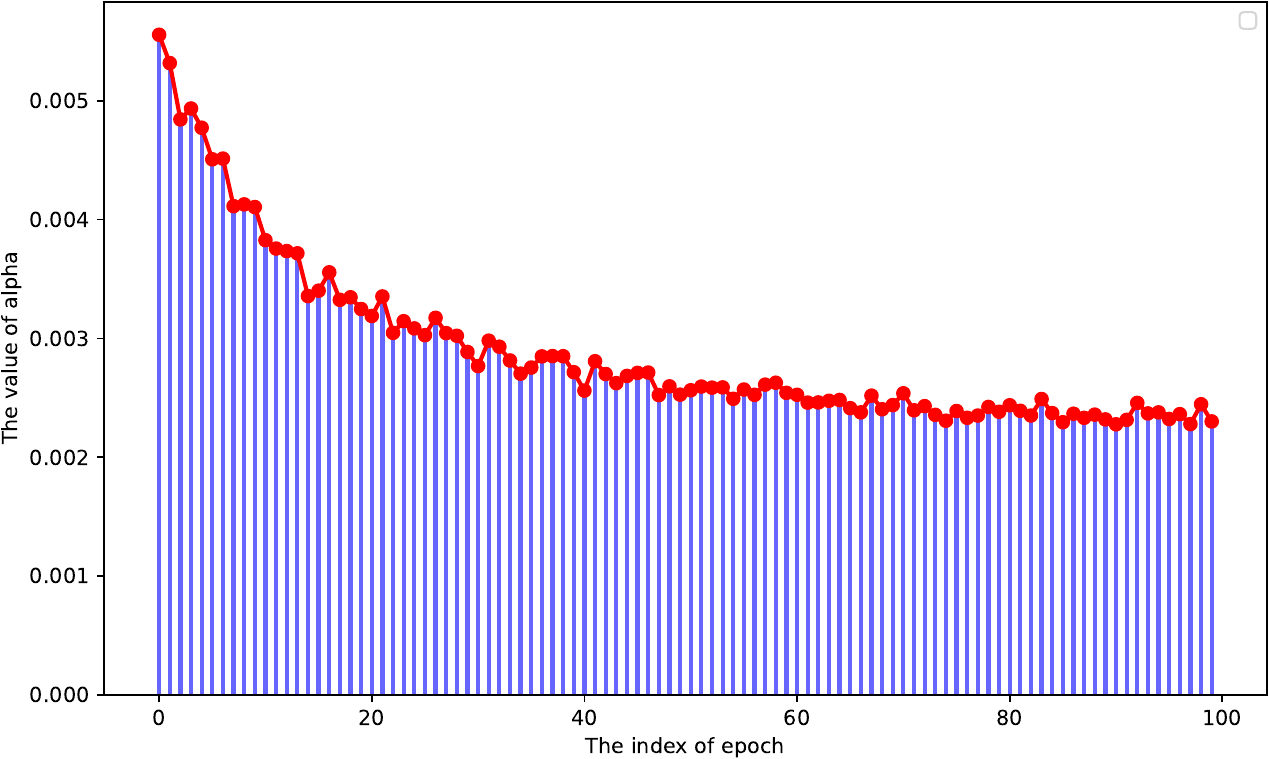}}
	\subfloat{\includegraphics[width=.25\linewidth]{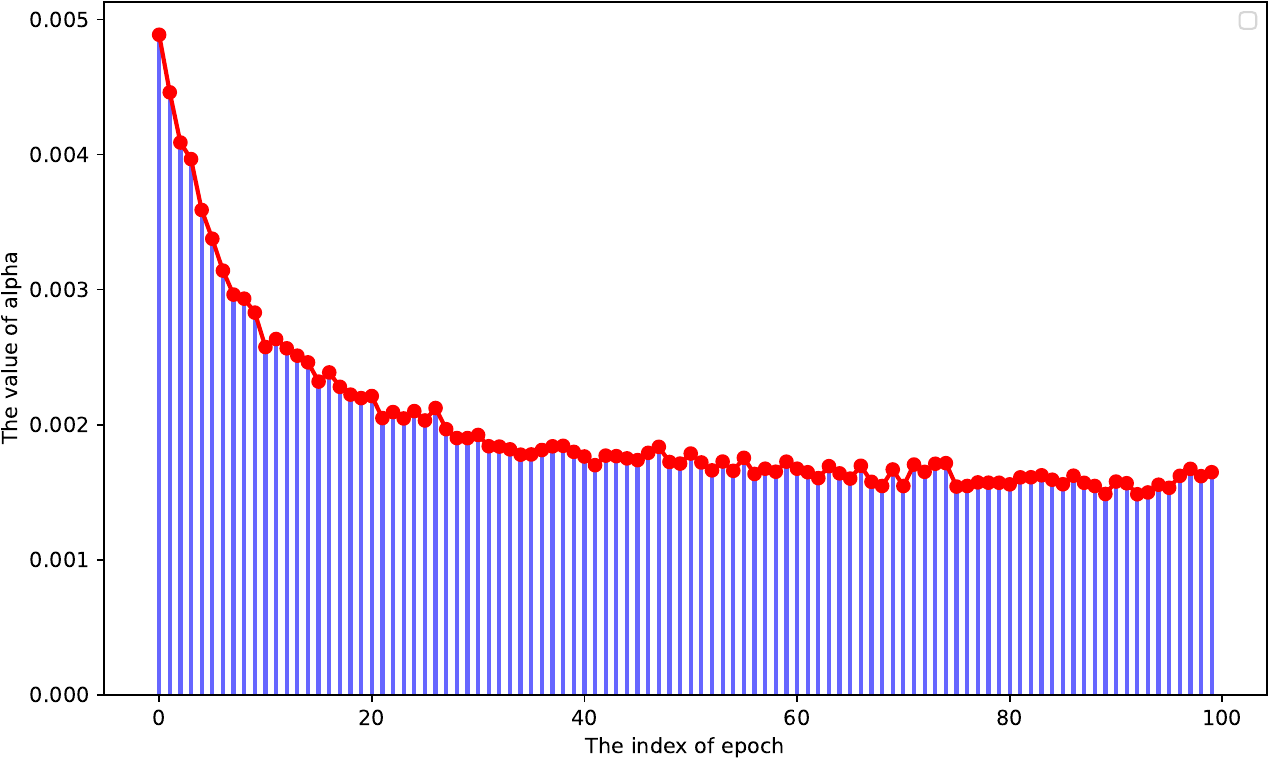}}
	\subfloat{\includegraphics[width=.25\linewidth]{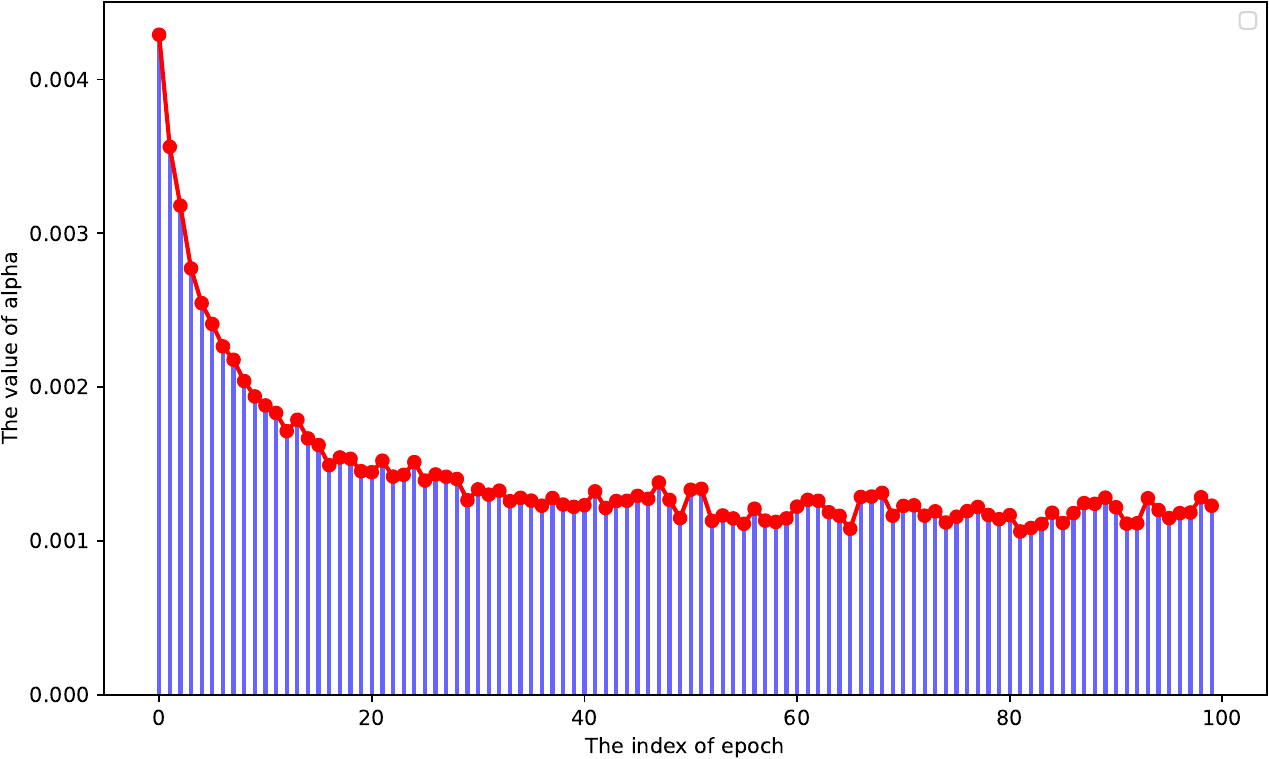}}
	\subfloat{\includegraphics[width=.25\linewidth]{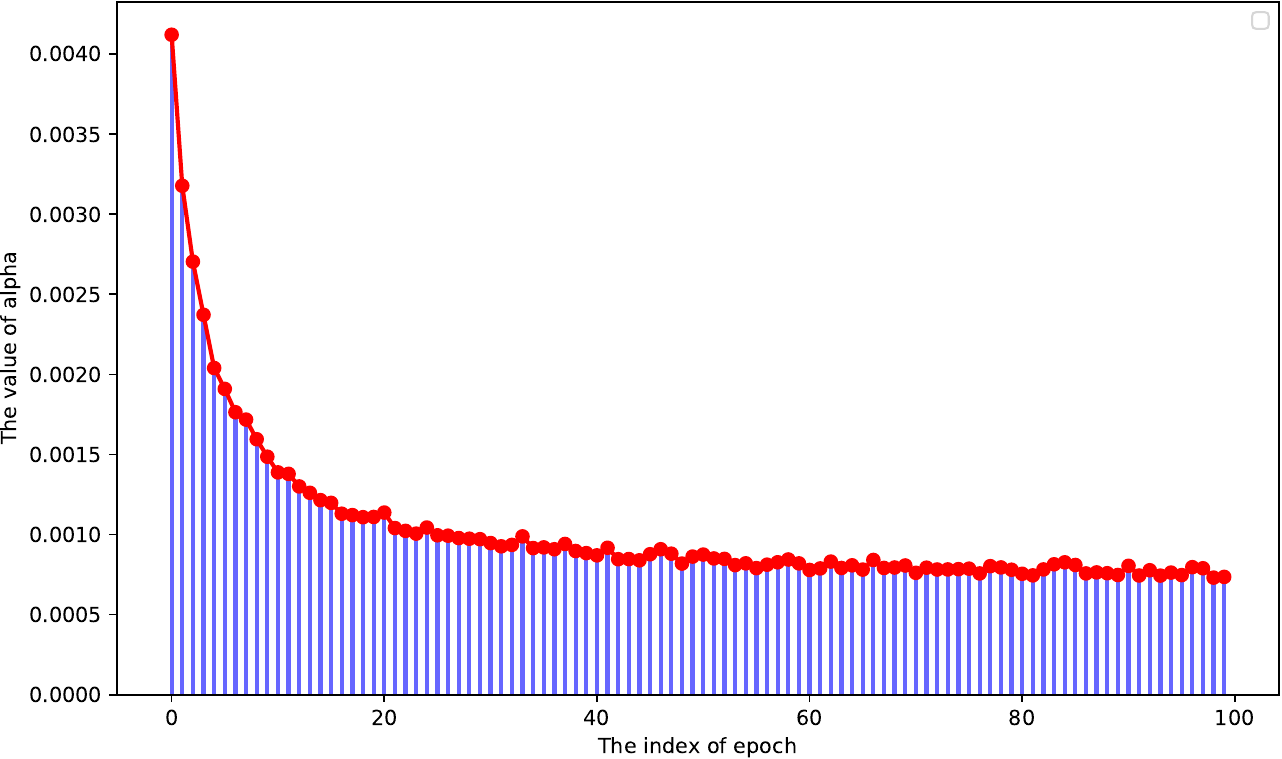}}\\
        \subfloat{\includegraphics[width=.25\linewidth]{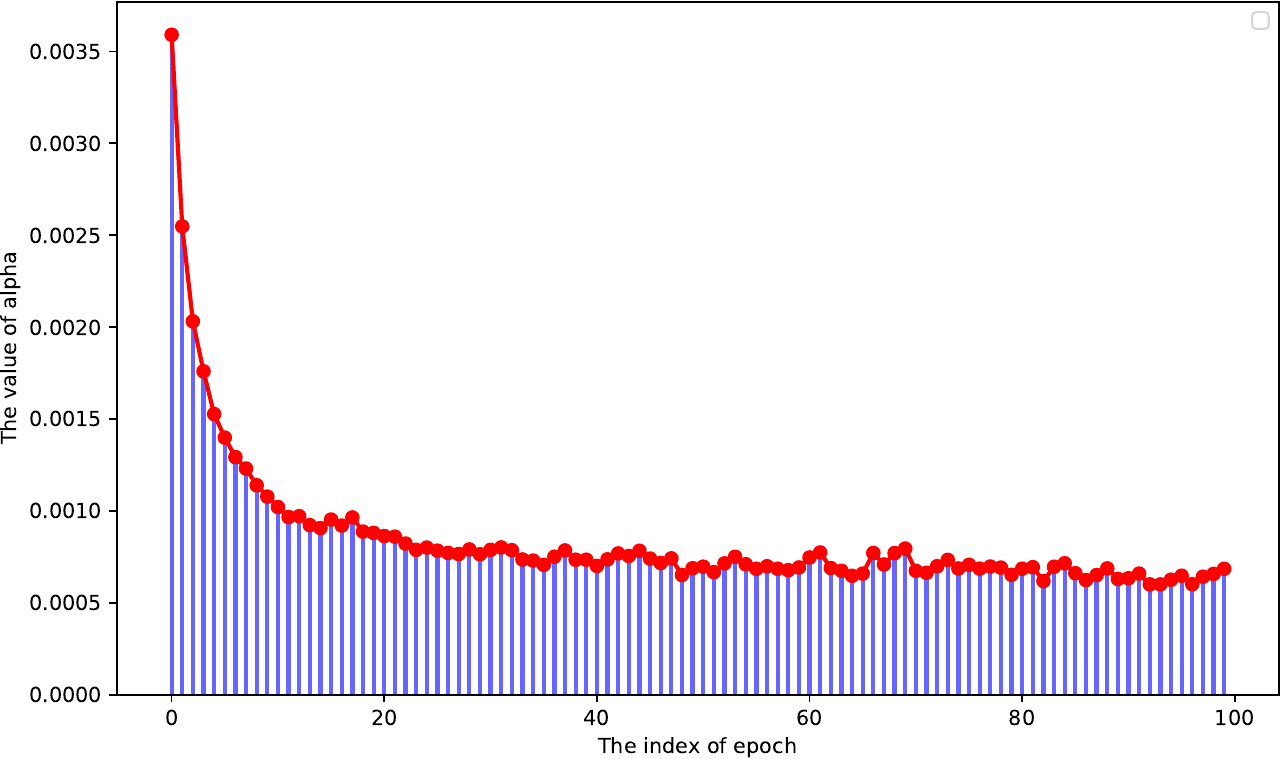}}
	\subfloat{\includegraphics[width=.25\linewidth]{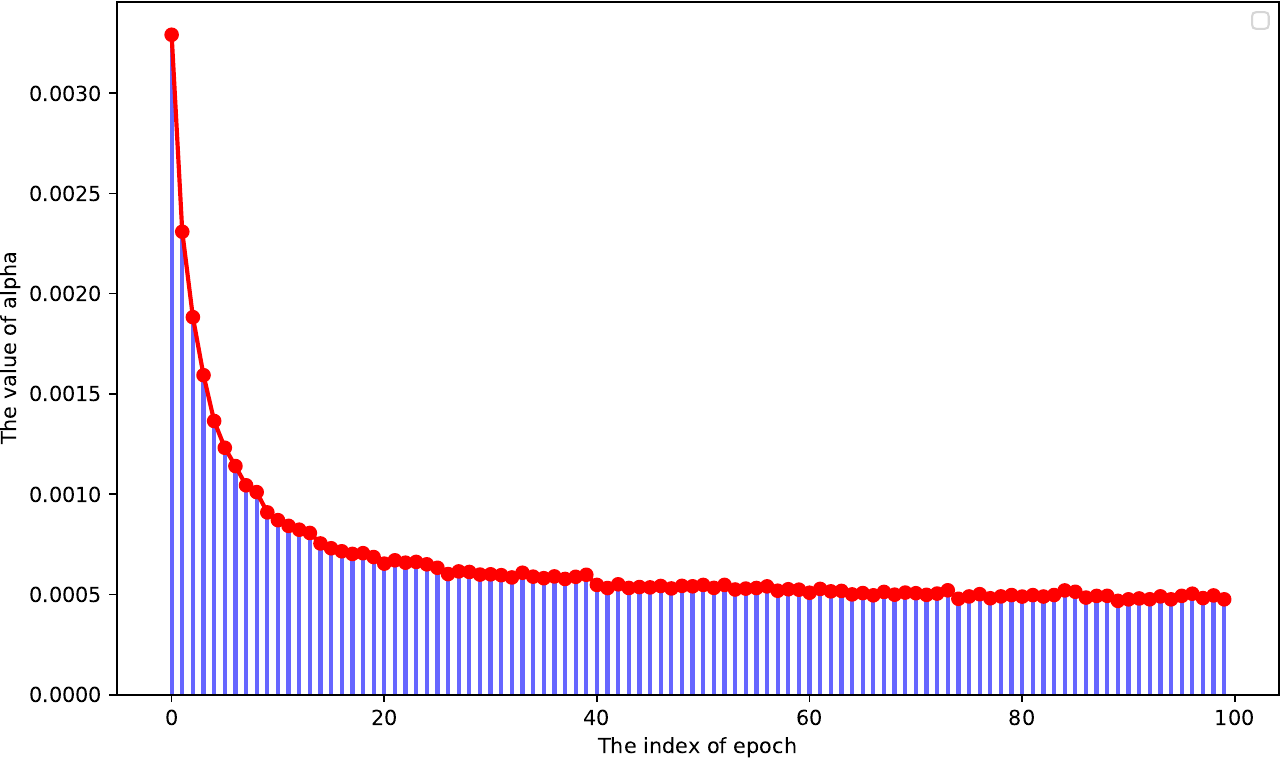}}
	\subfloat{\includegraphics[width=.25\linewidth]{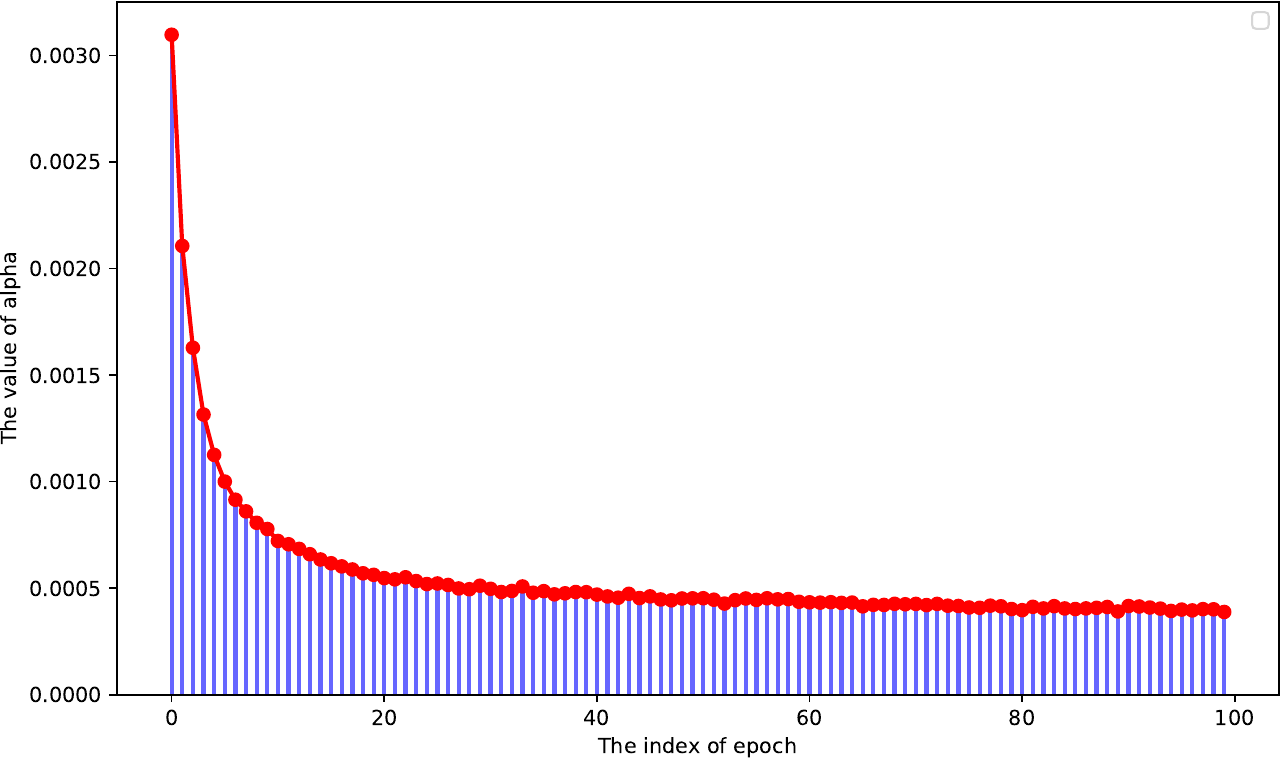}}
	\subfloat{\includegraphics[width=.25\linewidth]{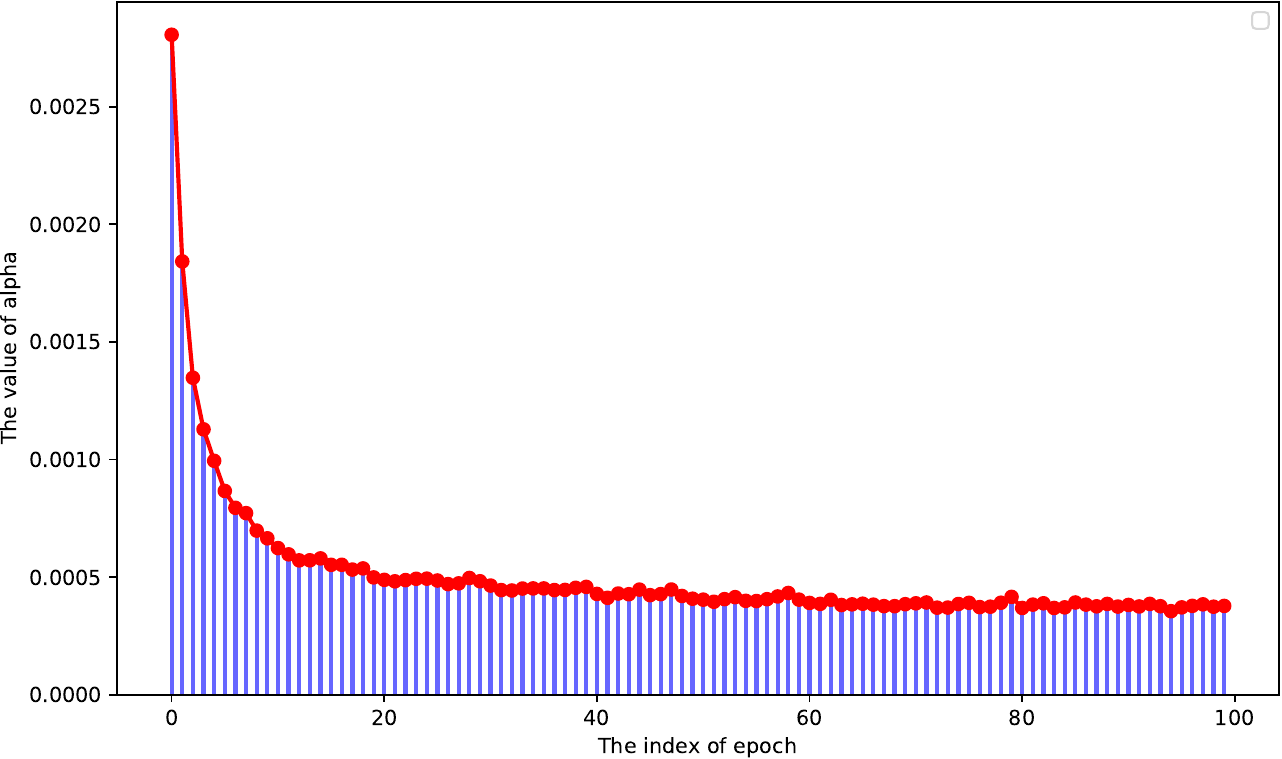}}\\
    
\caption{Distribution of $\alpha$ values under different communication rounds using data quantity-aware branch prediction by the client under different data amounts.}
    \label{fig_verify}
\end{figure*}

\subsection{Quantity-aware Branch}
If the client directly uploads the number of training data to the server, the server cannot verify its authenticity. To address this, we designed a quantity-aware branch and integrated it into the client model. This branch enables accurate prediction of the client's training data volume based on the model parameter change $\Delta\theta$, the learning rate $\eta$, and the average gradient of the mini-batch.

During the training of the neural network model, the training set is divided into multiple batches. The model is updated once for each batch, as shown in Equations \ref{one_ite_update} and \ref{one_epoch_update}, where $\nabla L_r(\theta_i)$ represents the model gradient at the $r$-th iteration. One epoch of training requires $R$ iteration updates. Therefore, the client's training data volume is $R \times \text{batch\_size}$. From Eq. \eqref{one_epoch_update}, we can deduce $R \approx \frac{\Delta \theta_i}{\eta \cdot \mathbb{E}[\nabla L(\theta_i)]}$, where $\Delta\theta_i$ is the model parameter difference for one epoch. However, since both $\Delta\theta_i$ and $\nabla L(\theta_i)$ are vectors rather than scalars, directly taking the norm for calculation may cause significant deviation. To address this, we use the quantity-aware branch to predict the adjustment factor $\alpha$. Using $\alpha$, we can accurately estimate the client's training data volume, as shown in Eq. \eqref{predict_T}.
\begin{align}\label{one_ite_update}
\theta^{'} = \theta - \eta \nabla L_r(\theta),
\end{align}
\begin{align}\label{one_epoch_update}
    \theta^{(t+1)} = \theta^t - \eta \sum_{r=1}^T \nabla L_r(\theta),
\end{align}
\begin{align}\label{predict_T}
R = \frac{\Delta \theta_i}{\eta \cdot \mathbb{E}[\nabla L(\theta_i)] \cdot \alpha}.
\end{align}

The client deploys the quantity-aware branch to predict the $\alpha$ value, as shown in Eq. \eqref{predict_alpha}. Here, $\varphi$ is the parameter of the quantity-aware branch, and $embedding(client_i)$ is the input to the quantity-aware branch corresponding to client $i$, which can either be fixed or learnable. The loss function of the quantity-aware branch is shown in Eq. \eqref{loss_prectict_alpha}. Using this loss function, we can train the quantity-aware branch model $f_{dua}$ so that its predicted $\alpha$ accurately estimates the client's training data volume.
\begin{align}\label{predict_alpha}
    \alpha = f_{dua}(\varphi; \text{embedding}(\text{client}_i)),
\end{align}
\begin{align}\label{loss_prectict_alpha}
    \text{$Loss$}_{\text{$dua$}} = \frac{1}{2} \left\| \frac{\Delta \theta_i}{\eta \cdot \mathbb{E}[\nabla L(\theta_i)] \cdot \alpha} - \left| D_i \right| \right\|^2.
\end{align}

The optimization process of the quantity-aware branch is as follows: the gradient of $Loss_{dua}$ with respect to $\alpha$ is shown in Eq. \eqref{Loss_pred/alpha}, and the chain rule for the derivative of the loss $Loss_{dua}$ with respect to $\varphi$ is shown in Eq. \eqref{Loss_pred/varphi}. Finally, the parameter $\varphi$ of the quantity-aware branch can be updated using Eq. \eqref{varphi_update}.
\begin{align}\label{Loss_pred/alpha}
    \begin{aligned}
        &\frac{\partial \text{$Loss$}_{\text{$dua$}}}{\partial \alpha}= \\ 
        & \left( \frac{\Delta \theta_i}{\eta \cdot \mathbb{E}[\nabla L(\theta_i)] \cdot \alpha} - \left| D_i \right| \right) \cdot \left( -\frac{\Delta \theta_i}{\eta \cdot \mathbb{E}[\nabla L(\theta_i)] \cdot \alpha^2} \right),
    \end{aligned}
\end{align}
\begin{align}\label{Loss_pred/varphi}
    \frac{\partial \text{Loss}_{\text{pred}}}{\partial \varphi} = \frac{\partial \text{Loss}_{\text{pred}}}{\partial \alpha} \cdot \frac{\partial \alpha}{\partial \varphi},
\end{align}
\begin{align}\label{varphi_update}
    \varphi^{(t+1)} = \varphi^{(t)} - \lambda \cdot \frac{\partial \text{Loss}_{\text{pred}}}{\partial \varphi}.
\end{align}

In FL, the client uploads the model gradient parameters to the server in each communication round, and the server aggregates the model gradients from the clients. During each communication round, the client typically performs $E$ epochs of training. After each epoch, the client predicts the adjustment factor $\alpha$ using the quantity-aware branch. At the end of the communication round, the client uploads the $E$ adjustment factors $\alpha$ and the parameter $\varphi$ of the quantity-aware branch to the server.

After receiving the model parameters uploaded by the client, the server will first verify whether the amount of training data reported by the client is accurate and reliable, based on the adjustment factor. For the specific verification process, refer to part \ref{Verify_Adjustment_Factor}. Once the server confirms that the reported data volume is accurate, it will perform weighted aggregation on the model parameters $\theta$ and the parameters $\varphi$ of the quantity-aware branch, based on the client's actual data volume.

Compared to directly calculating the adjustment factor $\alpha$ using Eq. \eqref{calculate_alpha}, using the quantity-aware branch to predict $\alpha$ offers three main advantages. First, it can handle the dynamic and complex characteristics of the client. Factors such as the distribution of client data, computing power, and training performance can all influence $\alpha$. The quantity-aware branch integrates these complex characteristics into the client’s embedding vector, allowing for a more accurate prediction of $\alpha$ and making the model more adaptable to the diversity of different clients. Second, it improves the robustness of the model. Direct calculation of $\alpha$ can lead to large deviations when there is noise or an attack during model training, which affects the aggregation quality of the global model. By learning the mapping from the embedding vector to $\alpha$, the model can tolerate noise or abnormal data, enhancing the system's robustness and fault tolerance. Third, the proposed method is highly scalable. The quantity-aware branch, as a module, can seamlessly integrate with other components of the model.
\begin{align}\label{calculate_alpha}
    \alpha = \frac{\Delta \theta_i}{\eta \cdot \mathbb{E}[\nabla L(\theta_i)] \cdot \left| D_i \right|}
\end{align}

\subsection{Reliability of the Adjustment Factor}
\label{Verify_Adjustment_Factor}
With its powerful computing capability, the server can calculate the trend of the adjustment factor $\alpha$ predicted by the quantity-aware branch under different training data volumes. Through experiments, we found that, under the same batch size, learning rate, and loss function, the $\alpha$ value of each communication round of the training model varies with different data volumes, as shown in Fig. \ref{fig_verify}. From Fig. \ref{fig_verify}, we observe significant differences in the size of the $\alpha$ value across communication rounds, but the distribution of the $\alpha$ value remains similar. Therefore, the server can acquire prior knowledge of the $\alpha$ value distribution through pre-training and use this information to verify whether the data volume reported by the client is accurate. If the reported data is found to be false, the server can issue an early warning to the client to discourage dishonest behavior.

Specifically, the server can predict the distribution of the predicted $\alpha$ values under different data volumes by fitting or using convolutional neural networks. The authenticity of the $\alpha$ value uploaded by the client can then be assessed based on the predicted distribution.

\begin{figure*}[t]
	\centering
	\subfloat{\includegraphics[width=.25\linewidth]{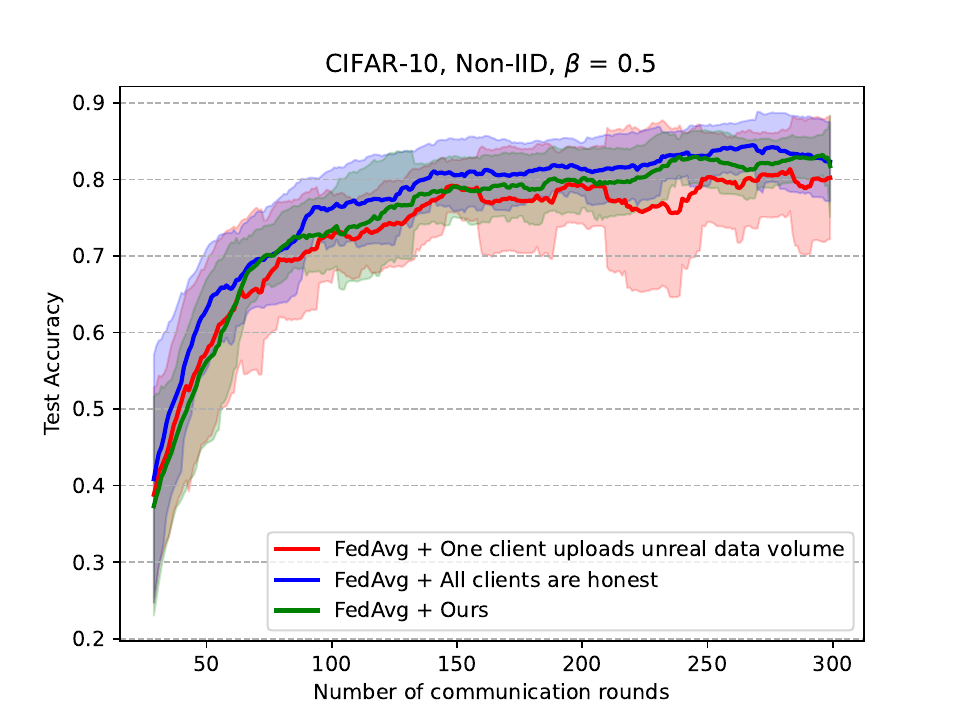}}
	\subfloat{\includegraphics[width=.25\linewidth]{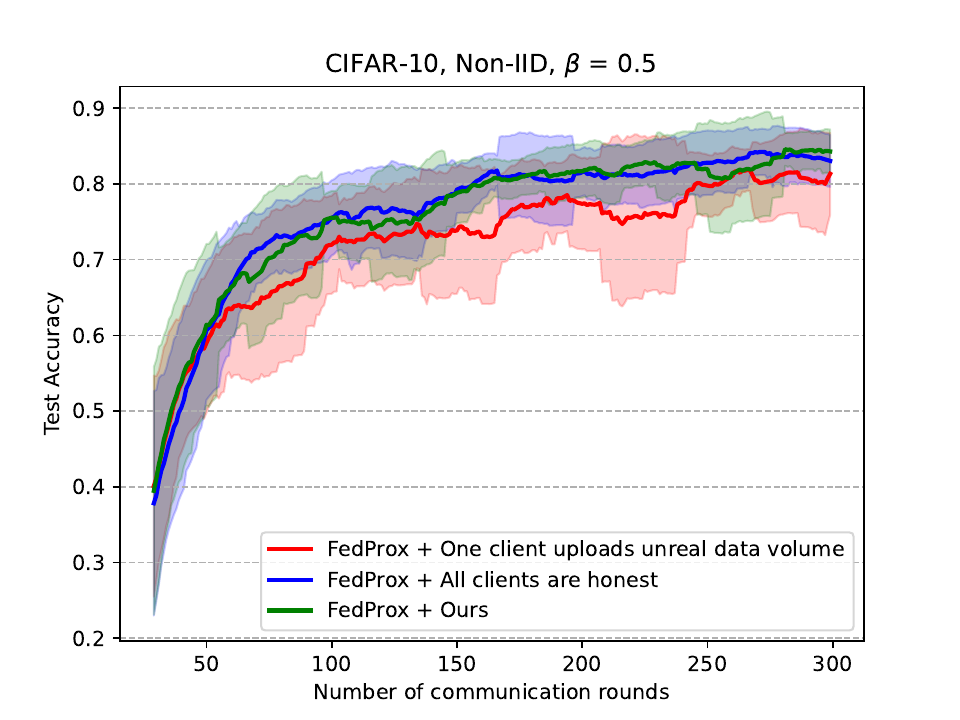}}
	\subfloat{\includegraphics[width=.25\linewidth]{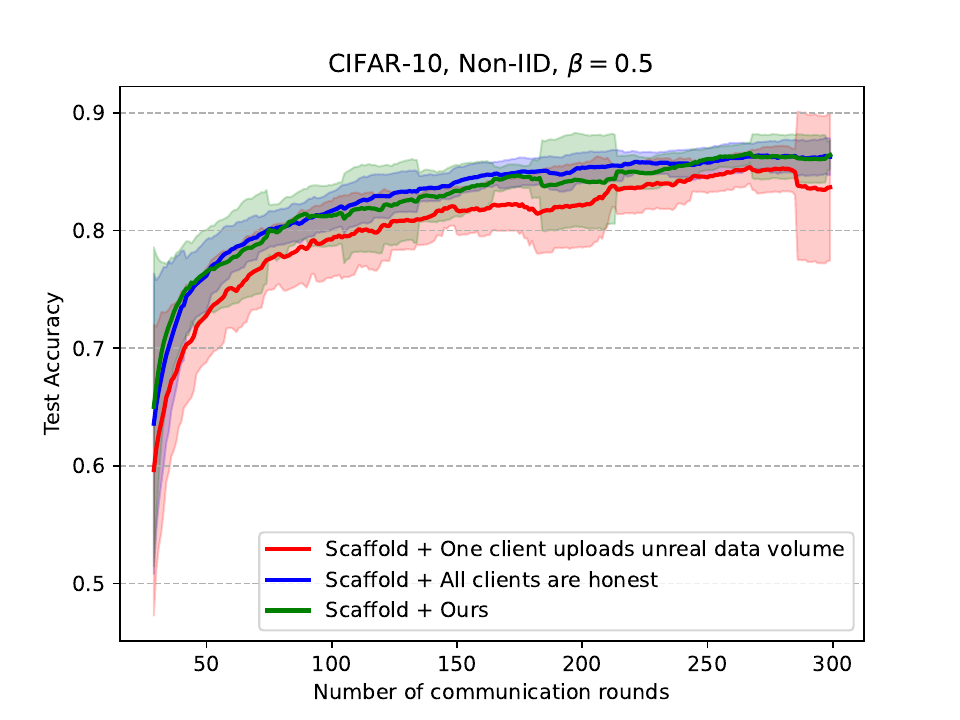}}
	\subfloat{\includegraphics[width=.25\linewidth]{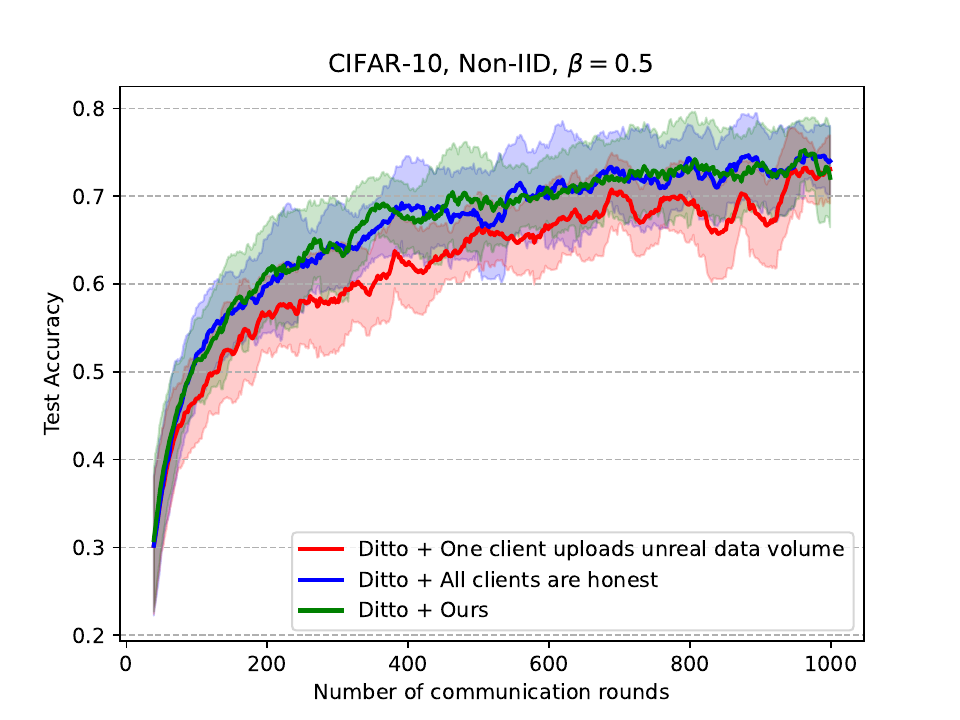}}\\
    (a) CIFAR-10
    \subfloat{\includegraphics[width=.25\linewidth]{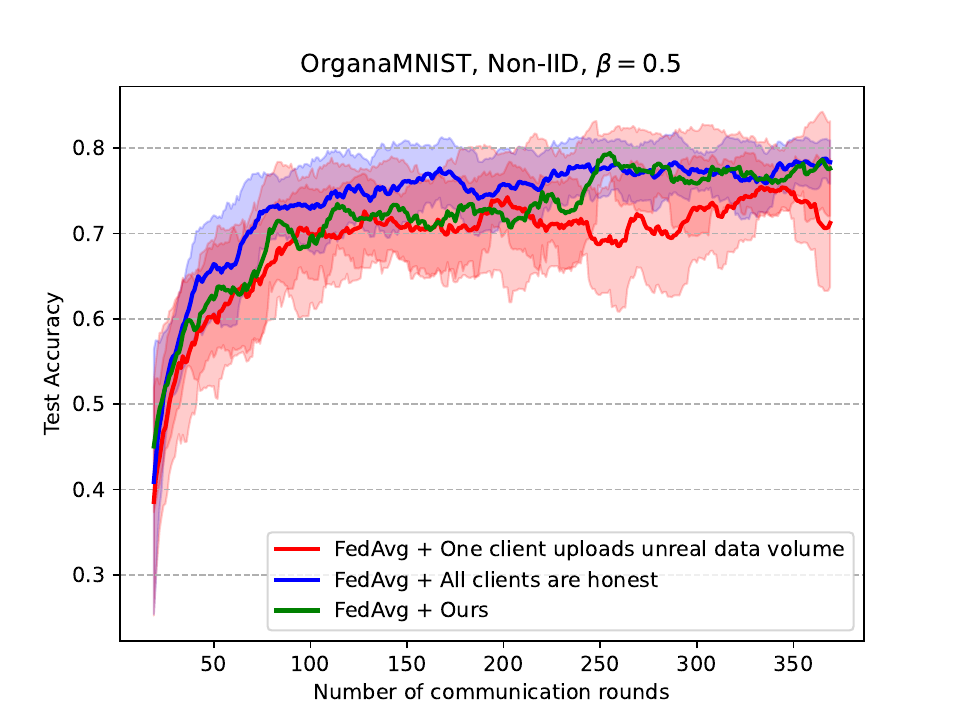}}
	\subfloat{\includegraphics[width=.25\linewidth]{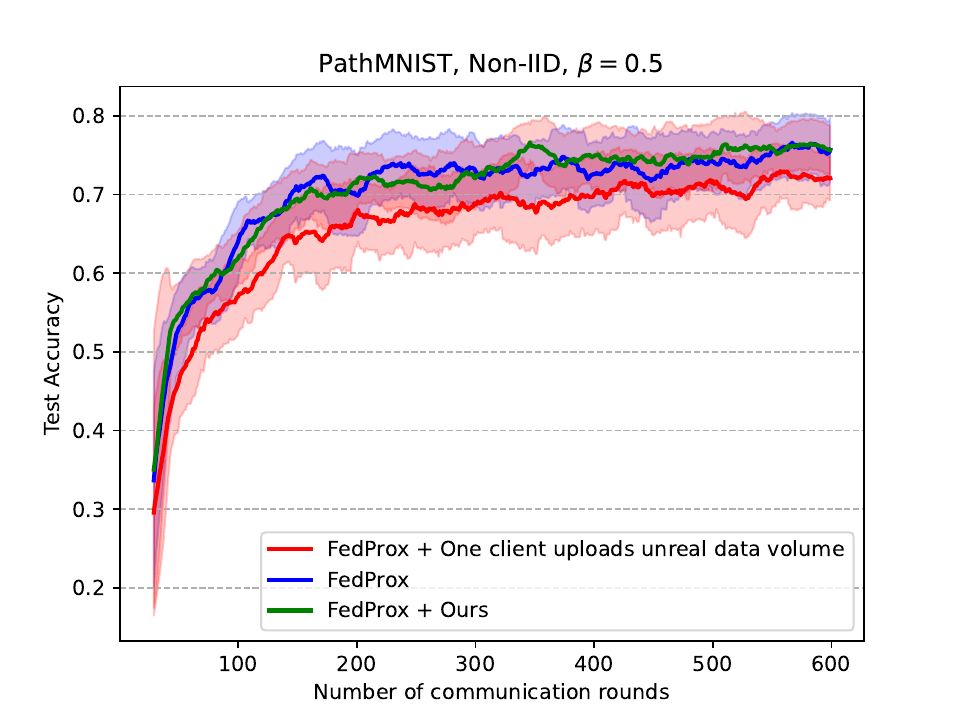}}
	\subfloat{\includegraphics[width=.25\linewidth]{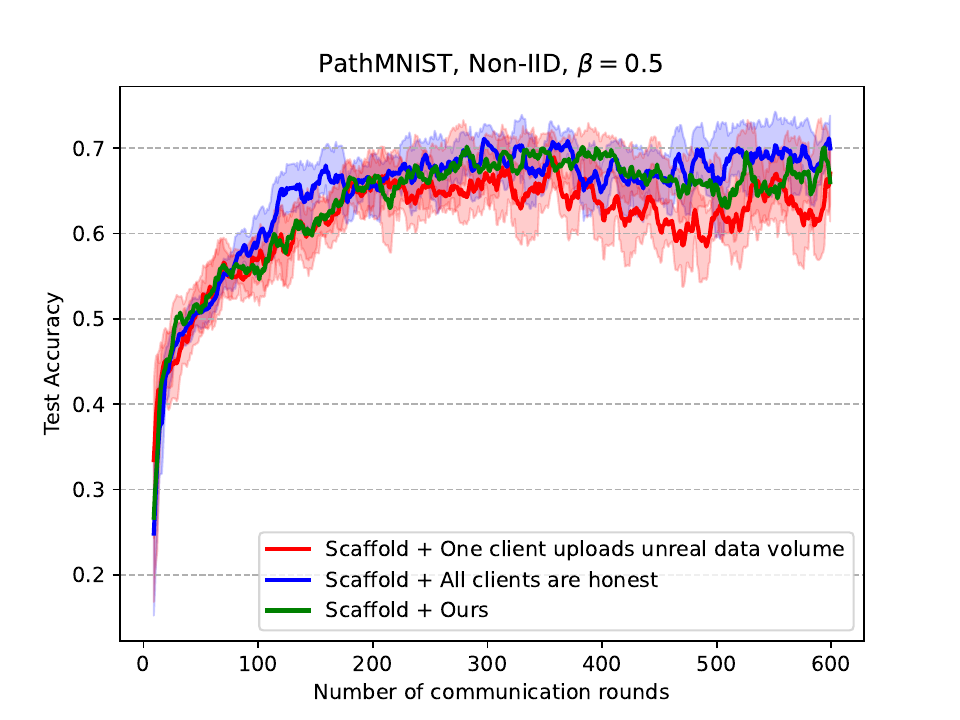}}
	\subfloat{\includegraphics[width=.25\linewidth]{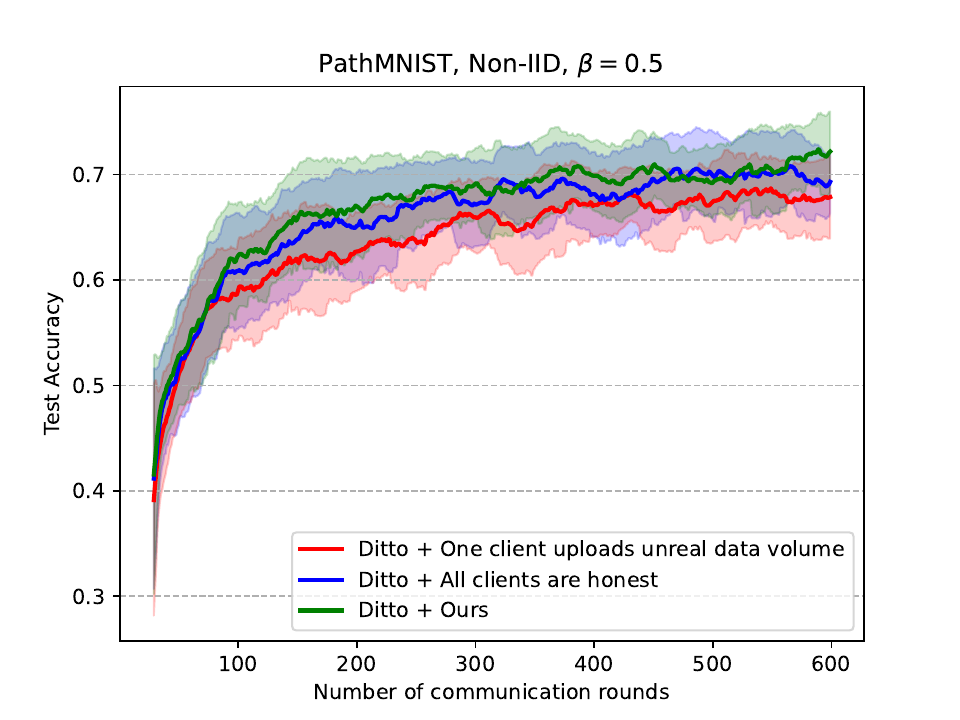}}\\
    (b) MedMNIST-PathMNIST
    \subfloat{\includegraphics[width=.25\linewidth]{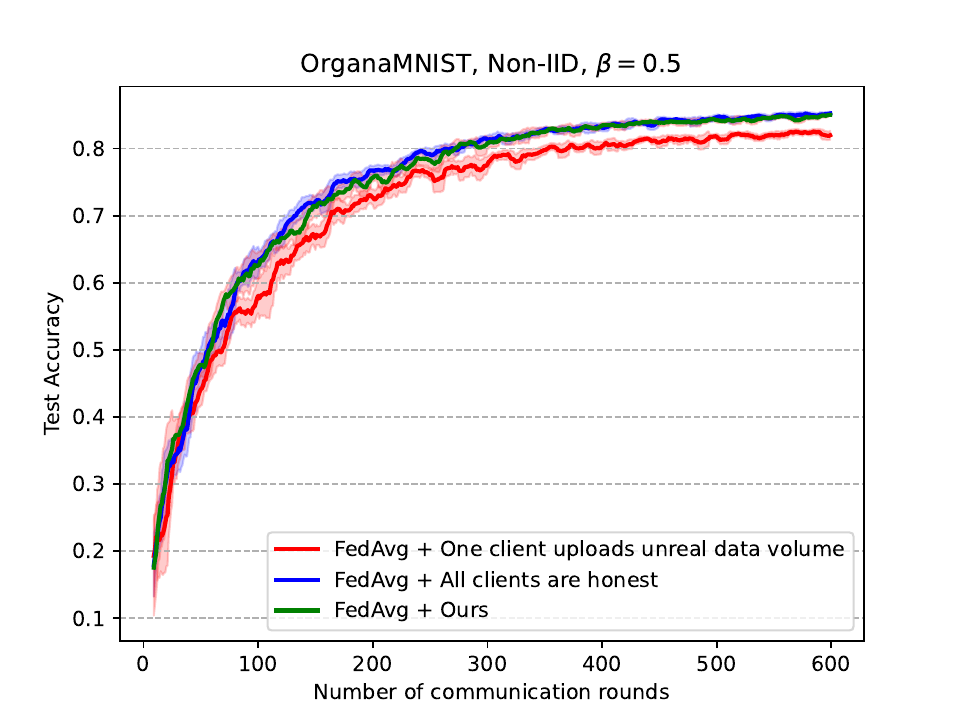}}
	\subfloat{\includegraphics[width=.25\linewidth]{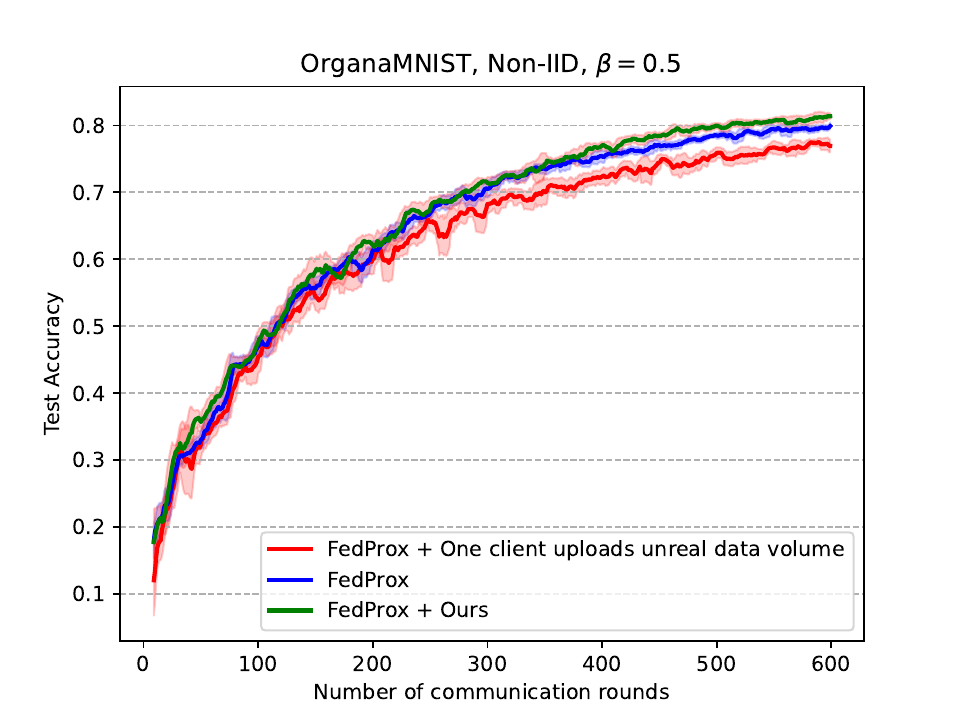}}
	\subfloat{\includegraphics[width=.25\linewidth]{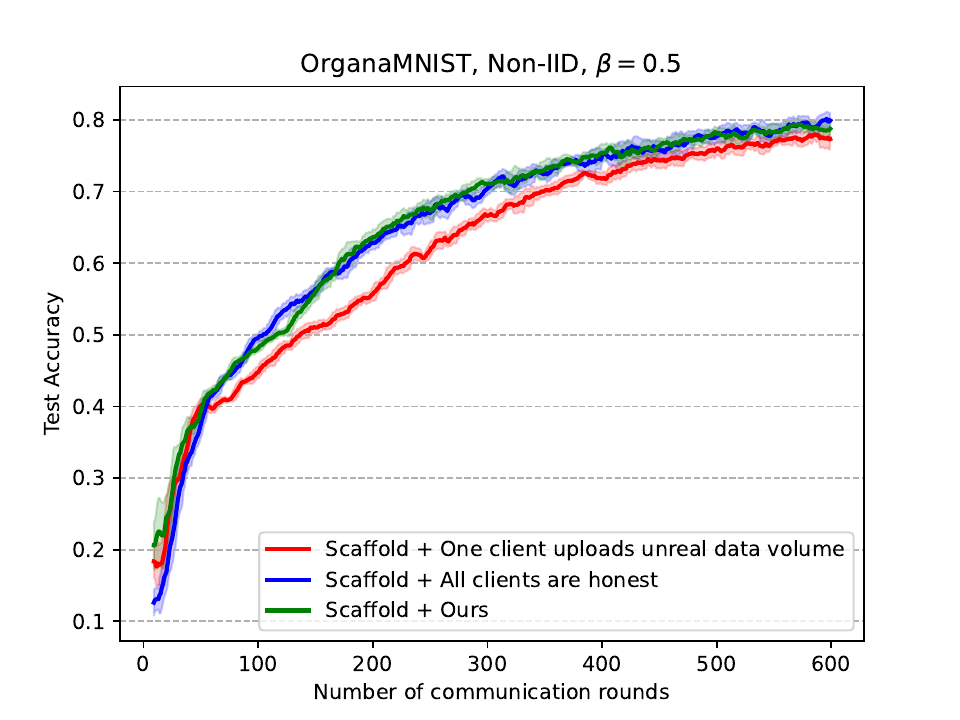}}
	\subfloat{\includegraphics[width=.25\linewidth]{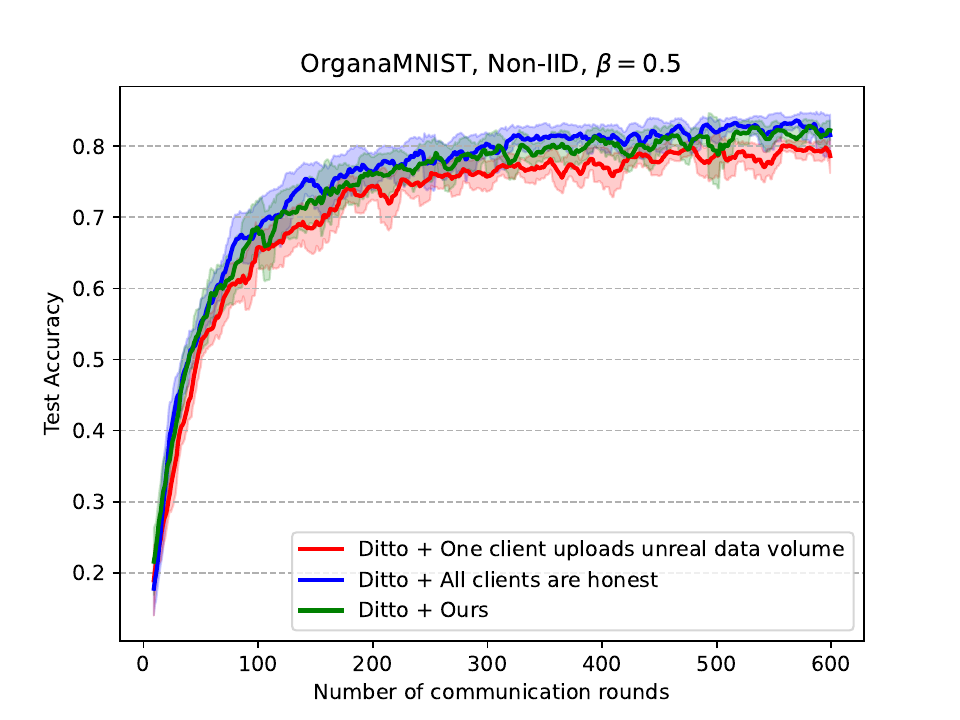}}\\
    (c) MedMNIST-OrganaMNIST
	\caption{The proposed algorithm is tested with four FL methods: FedAvg, FedProx, Scaffold, and Ditto. Two scenarios are considered: (1) all clients behave honestly, and (2) one client uploads falsified data volumes. The evaluation measures test accuracy on the CIFAR-10, MedMNIST-PathMNIST, and MedMNIST-OrganaMNIST datasets. The datasets are non-IID, with a Dirichlet distribution ($\beta=0.5$) used to define the data distribution.}
    \label{exp_results_1}
\end{figure*}

\section{EXPERIMENTAL EVALUATION}

\subsection{Experiment Setup}
In the experiment, we set the number of clients to 100 and selected 10 clients in each round to participate in federated training. One client was set to report three times the amount of data to the server in each training round, simulating a scenario with a dishonest client. Other experimental settings are as follows.

\textbf{Datasets}. We selected CIFAR-10 \cite{krizhevsky2009learning} and the medical dataset MedMNIST for the experiments. CIFAR-10 is an image dataset with 10 categories, containing 50,000 training samples and 10,000 test samples, all of which are 3-channel images of size 32*32. MedMNIST \cite{yang2023medmnist} is a medical dataset that includes 12 2D sub-datasets and 6 3D sub-datasets, with image types covering X-rays, CT scans, MRIs, and more. For this experiment, we selected two 2D sub-datasets from MedMNIST: OrganaMNIST \cite{bilic2023liver} and PathMNIST \cite{kather2019predicting}.

\textbf{Comparison algorithms}. In this paper, we selected four classic FL algorithms: FedAvg, FedProx, Ditto, and Scaffold. Among them, FedAvg is the most fundamental FL algorithm and introduced the concept of FL. The FedProx algorithm adds a regularization term to FedAvg to address the heterogeneity problem in FL. The Ditto algorithm deploys both personalized and global models on the client, using the global model as a constraint on the personalized model during training to prevent the personalized model from deviating too far from the global model. The Scaffold algorithm introduces control variables and control gradients, ensuring model stability even in scenarios with heterogeneous data.

\subsection{Results and Discussion}

We evaluate four FL algorithms—FedAvg \cite{mcmahan2017communication}, FedProx \cite{li2020federated}, Scaffold \cite{karimireddy2020scaffold}, and Ditto \cite{li2021ditto}—on the three datasets mentioned above to compare the accuracy of the proposed algorithm under two scenarios: when all clients are honest and when one client uploads unreal data volume. In the experiment, we use a Dirichlet distribution to partition the Non-IID data for each client, with $\beta = 0.5$. The experimental results are shown in Fig. \ref{exp_results_1}.

From Fig. \ref{exp_results_1}, we can observe that when a client uploads an unreal amount of training data during FL, it negatively affects the performance of the aggregated global model. For example, in the Cifar-10 dataset, when a client reports an unreal amount of training data, the accuracy of the global model after aggregation decreases by 2.58\%, 3.79\%, 3.2\%, and 3.51\% for the four FL algorithms—FedAvg, FedProx, Scaffold, and Ditto—compared to the weighted aggregation based on the client's actual data volume. Furthermore, we can see that when the client uploads unreal data, the accuracy of the global model fluctuates significantly during training. This is because unreal reporting of client data volume leads to inaccurate weighted aggregation, which over-amplifies the influence of dishonest clients. In Non-IID settings, clients that report unreal data volumes may provide data that differs significantly from the global distribution, causing the global model to deviate from the true characteristics of the overall data, thus reducing accuracy. During training, the model updates become unstable due to the unreal data volume, and the imbalance in weights leads to frequent adjustments in the global model, resulting in large accuracy fluctuations.

Similarly, when clients dishonestly report the amount of training data, the accuracy of the global model using the proposed method, \methodname{}, remains nearly the same as when the data is honestly reported. This is because, in the \methodname{} algorithm, the server can verify the authenticity of the training data uploaded by each client based on the distribution of the adjustment factor $\alpha$, combined with the prior knowledge of $\alpha$ distribution obtained through pre-training. If a dishonest client is detected, the server will issue a warning. If the client continues to upload false data volume, it will be excluded from federated training.

\textbf{Computation and communication complexity analysis}. The proposed FedDua algorithm introduces two additional computations: client training of the data quantity-aware branch and server pre-training of prior knowledge about the $\alpha$ distribution. The server possesses sufficient computing, storage, and communication capabilities. Therefore, the additional computational cost primarily arises from client training of the data quantity-aware branch. Typically, the number of parameters in the data quantity-aware branch model, $\varphi$, is much smaller than the local model $\theta$ on the client, and the corresponding computational complexity is $O(\frac{1}{d}\theta)$, with $d>10$. As a result, the additional computational cost for the client due to the data quantity-aware branch is less than 10\%. Furthermore, no additional communication cost is incurred.


\section{CONCLUSIONS}
In the proposed \methodname{} algorithm, the data quantity-aware branch and server pre-training enable accurate judgment of the authenticity of the data uploaded by the client and prediction of the client's data volume. The proposed method addresses the issue of performance degradation in the global model due to clients dishonestly reporting their data volume during weighted aggregation in FL. However, this paper focuses only on dishonestly reported data volume. The quality of the data also significantly impacts the performance of the aggregated global model. In future work, we will analyze weighted aggregation in FL from the perspective of data quality to further improve model performance.




\section{ACKNOWLEDGMENTS}

This work is funded in part by an international Collaboration Fund for Creative Research of National Science Foundation of China (NSFC lCFCRT) under the Grant no. W2441019; National Natural Science Foundation of China (62136003); Shanghai Pujiang Program (22PJ1423400) and Shanghai Sailing Program (22YF1401300); the Ministry of Education, Singapore, under its Academic Research Fund Tier 1; the National Research Foundation, Singapore and DSO National Laboratories under the AI Singapore Programme (AISG Award No. AISG2-RP-2020-019).

\bibliographystyle{IEEEbib}
\bibliography{main}

\end{document}